%% file: coling_latex.tex
\pdfoutput=1

\documentclass[11pt]{article}

\usepackage[final]{coling}

\usepackage{times}
\usepackage{latexsym}

\usepackage[T1]{fontenc}

\usepackage[utf8]{inputenc}

\usepackage{microtype}

\usepackage{inconsolata}

\usepackage{arydshln}

\usepackage{algorithm} 
\usepackage{algpseudocode}

\usepackage[skins]{tcolorbox}
\usepackage{graphicx}
\usepackage{enumitem}
\usepackage{makecell}

\tcbuselibrary{listingsutf8}
\global

\tcbset{
  aibox/.style={
    width=474.18663pt,
    top=10pt,
    colback=white,
    colframe=black,
    colbacktitle=black,
    enhanced,
    center,
    attach boxed title to top left={yshift=-0.1in,xshift=0.15in},
    boxed title style={boxrule=0pt,colframe=white,},
  }
}
\newtcolorbox{AIbox}[2][]{aibox,title=#2,#1}

\usepackage{amsmath}

\usepackage{graphicx}
\usepackage{cleveref}
\usepackage{booktabs}
\usepackage{subcaption}
\usepackage{array}

\usepackage{soul}

\usepackage{caption}
\usepackage{subcaption}

\usepackage{booktabs}
\usepackage{multirow}
\usepackage{makecell}

\usepackage{enumitem}

\usepackage{annotation}
\usepackage{tikz}
\usepackage{hyperref}

\newcommand{\metric}{MESA}

\title{Is my Meeting Summary Good? \\ Estimating Quality with a Multi-LLM Evaluator}

\author{Frederic Kirstein\textsuperscript{1,*}, Terry Ruas\textsuperscript{1}, Bela Gipp\textsuperscript{1} \\
  \textsuperscript{1}University of Göttingen, Germany \\
\textsuperscript{*}\texttt{kirstein@gipplab.org} }

\begin{document}
\maketitle
\AddAnnotationRef

\begin{abstract}
The quality of meeting summaries generated by natural language generation (NLG) systems is hard to measure automatically.
Established metrics such as ROUGE and BERTScore have a relatively low correlation with human judgments and fail to capture nuanced errors.
Recent studies suggest using large language models (LLMs), which have the benefit of better context understanding and adaption of error definitions without training on a large number of human preference judgments.
However, current LLM-based evaluators risk masking errors and can only serve as a weak proxy, leaving human evaluation the gold standard despite being costly and hard to compare across studies.
In this work, we present \metric{}, an LLM-based framework employing a three-step assessment of individual error types, multi-agent discussion for decision refinement, and feedback-based self-training to refine error definition understanding and alignment with human judgment.
We show that \metric{}'s components enable thorough error detection, consistent rating, and adaptability to custom error guidelines.
Using GPT-4o as its backbone, \metric{} achieves mid to high Point-Biserial correlation with human judgment in error detection and mid Spearman and Kendall correlation in reflecting error impact on summary quality, on average 0.25 higher than previous methods.
The framework's flexibility in adapting to custom error guidelines makes it suitable for various tasks with limited human-labeled data.
\end{abstract}

\section{Introduction}
\input{text/01_Introduction}

\section{Methodology}
\input{text/03_Methodology}

\section{Experiments}
\input{text/04_Experiments}

\section{Related Work}
\input{text/02_Related_Work}

\section{Final Considerations}
In this paper, we introduced \metric{}, an LLM-based single-aspect evaluation framework for meeting summarization using a three-step evaluation pipeline and multi-agent discussion paradigm.
We conducted extensive experiments on the influence of the individual components and assessment performance of the framework using a modified version of the QMSum Mistake dataset annotated by humans on eight error types.
Experiments revealed that \metric{} identifies error instances more thoroughly and better captures impact than established metrics, achieving a higher correlation with human judgment.
The self-training approach enhances alignment with human assessments and reduces oversensitive detections.
The framework's flexibility in allowing for custom error guidelines and adapting to human scoring behavior with minimal samples makes it applicable beyond meeting summarization for tasks with similar limitations.
We will release the codebase and updated dataset to encourage research on LLM-based evaluation.

\section*{Acknowledgements}
This work was supported by the Lower Saxony Ministry of Science and Culture and the VW Foundation.
Frederic Kirstein was supported by the Mercedes-Benz AG Research and Development.

\section*{Limitations}
We have used large LLMs in this work (GPT4) and have not explicitly studied whether the approach works on smaller models.
As we used smaller models while exploring multi-agent discussions, we could observe a similar level of detail generated by the smaller models.
This observation indicates that the approach can also be successful with models from the 10B to 30B parameter category.

Another possible weakness of our work could be that we carry our experiments on a dataset that might seem small (i.e., 170 samples).
However, its size is comparable to that of the original, established QMSum dataset (232 samples) and the original QMSum Mistake dataset (200 samples).
We contribute to refining the original datasets by carefully annotating human errors, curating reasoning traces, and defining new error types.
As there are no large, high-quality datasets available with diverse meeting types due to data security and intellectual property constraints, a method to generate synthetic meetings on a human-like level would be required to mitigate this data scarcity.

Further, we only investigate and report metric performance measured as accuracy or correlation, leaving out computational requirement concerns.
We do so as the LLM-based approaches will be more costly than the established count-based and model-based metrics.
We include in our experiments a more lightweight version of \metric{} to demonstrate that a weaker, less expensive variant yields similar results as our best-performing option.

\section*{Ethics Statement}
\paragraph{Licenses:}
We adhered to licensing requirements for all tools used (OpenAI, Microsoft, Google, Meta, Huggingface).

\paragraph{Privacy:}
User privacy was protected by screening the dataset for personally identifiable information during quality assessment.

\paragraph{Intended Use:}
Our pipelines are intended for organizations to quickly and efficiently assess the quality of summaries and extend their summarization systems with a feedback-generating mid-layer.
While poor summary quality assessment may affect user experience and the performance of depending systems, it should not raise ethical concerns as the evaluation is based solely on given transcripts and summaries.
Production LLMs will only perform inference, not re-training on live transcripts.
Assessments will be accessible only to meeting participants, ensuring information from other meetings remains confidential.

\bibliography{25_COLING_MESA}

\appendix


\section{Prompts}
\input{text/B_Appendix_Prompts}

\section{Example Outputs}
\input{text/C_Appendix_ExampleOutputs}

\section{Error Types}
\input{text/D_Appendix_ErrorTypeDefinitions}

\section{Dataset}
\input{text/E_Appendix_Dataset}

\section{Balanced Accuracy Definition}
\input{text/F_Appendix_BACC}

\clearpage
\onecolumn
\input{annotation.tex}

\end{document}

%% file: text/01_Introduction.tex
\input{figures/main_figure}

Meeting summaries have become integral to professional environments \cite{ZhongYYZ21g, HuGDD23a, LaskarFCB23b}, serving as references, updates for absentees, and reinforcements of key topics discussed.
The integration of summarization services into established digital meeting platforms (e.g., Zoom\footnote{\href{https://www.zoom.com/en/ai-assistant}{https://www.zoom.com/en/ai-assistant}},
Microsoft Teams\footnote{\href{https://copilot.cloud.microsoft}{https://copilot.cloud.microsoft}}, 
Google Meet\footnote{\href{https://support.google.com/meet/answer/14754931}{https://support.google.com/meet/}}) further underscores their growing relevance.
The evaluation of generated summaries remains an ongoing problem \cite{KirsteinWGR24c} and is typically solved through costly, time-consuming human assessment.
Consequently, an automatic evaluator is necessary, which would, if providing insights along the scoring, also enable sophisticated techniques such as feedback-based summary refinement \cite{KirsteinRG24a} and reinforcement learning from AI feedback \cite{LeePMM23}.

Established automatic metrics such as ROUGE \cite{Lin04}, BERTScore \cite{ZhangKWW20}, and BARTScore \cite{YuanNL21} exhibit a relatively low correlation with human judgment.
These count- and model-based metrics often fail to reliably detect errors, leading to error masking \cite{KirsteinWRG24b}, and lack sensitivity to error impact, resulting in inaccurate reflection of summary quality in score \cite{KirsteinRG24a}.

Recently, Large language models (LLMs) have been proposed as evaluators for text summarization \cite{LiuIXW23,LiuYHZ23,WangKGY24}, assigning Likert scores based on predefined guidelines.
However, these approaches face limitations in meeting summarization contexts.
Current annotation guidelines do not cover typical errors in meeting summaries, e.g., structure presentation, coreference issues \cite{KirsteinWRG24b}, resulting in oversight and insufficient quality assessment.
Moreover, the subjective nature of existing guidelines, e.g., 'informativeness' \cite{LiuYHZ23} may lead to inconsistent interpretations by LLMs, resulting in unreliable evaluations \cite{KirsteinRG24a}.

We introduce the meeting summary assessor (\metric{}), a multi-stage LLM-based framework that mimics the human evaluation approach (see \Cref{fig:main_figure}).
\metric{} operates on three levels: error-specific evaluation, overall evaluation, and self-training.
For each error type to be considered, an \textbf{error-specific evaluation} is performed that employs a three-step process to identify potential errors, assess their impact, and assign Likert scores (0-5) \cite{Likert32}, utilizing chain-of-thought (CoT) prompting \cite{WeiWSB24} and verbose confidence scores (0-10) \cite{TianMZS23} to boost performance.
The three-step assessment can be carried out using a multi-agent discussion protocol \cite{LiangHJW23} where one agent generates a draft challenged and refined by other agents, allowing for a dynamic refinement step considering different perspectives \cite{LiZWH24}.
The \textbf{overall evaluation} synthesizes the individual Likert scores into an overall rating of the error impact (0-5) and a corresponding quality score (1-10).
The \textbf{self-training} mechanism, inspired by \citet{WangKGY24}'s self-teaching and \citet{KirsteinRG24a}'s feedback approach, influences the evaluation behavior by comparing \metric{}'s assessments with available human annotations.
We employ an LLM judge \cite{ZhengCSW24} to evaluate reasoning quality and predefined categories for labeling Likert score discrepancies.
The comparisons are processed by a second LLM that generates a feedback report pointing out how \metric{} should change behavior to better align with human judgment in scoring and reasoning.
This feedback is appended to the prompts of the error-specific evaluation.

We evaluate \metric{} using available error definitions and a modified version of QMSum Mistake \cite{KirsteinRG24a}, combining total and partial omission errors.
Experiments with GPT-4o\footnote{We will refer to this as GPT4 throughout the paper.} as the backbone model demonstrate \metric{}'s strong performance across all error types, outperforming existing evaluators in error existence correlation (avg. gap: $\sim$0.2) and severity representation (avg. gap: $\sim$0.25).
We observe that the self-training step helps align with human judgment, mitigating overly harsh scoring tendencies and reducing the false-positive detection of error instances.
The three-step error-specific evaluation allows for a thorough analysis, reducing false-negative detection.
Our contributions are summarized as follows:
\begin{itemize}[parsep=0.3pt,itemsep=0.1pt]
    \item A multi-agent-based, self-training evaluation framework, \metric{}, that outperforms baseline metrics on meeting summary assessment.
    \item A thorough analysis of the components (i.e., three-step evaluation, single-aspect processing, multi-agent discussion, self-training).
    \item We introduce multi-agent discussion to the meeting summarization domain and propose a three-step evaluation to boost performance.
\end{itemize}

%% file: figures/main_figure.tex
\begin{figure*}
    \centering
    \includegraphics[width=1\linewidth]{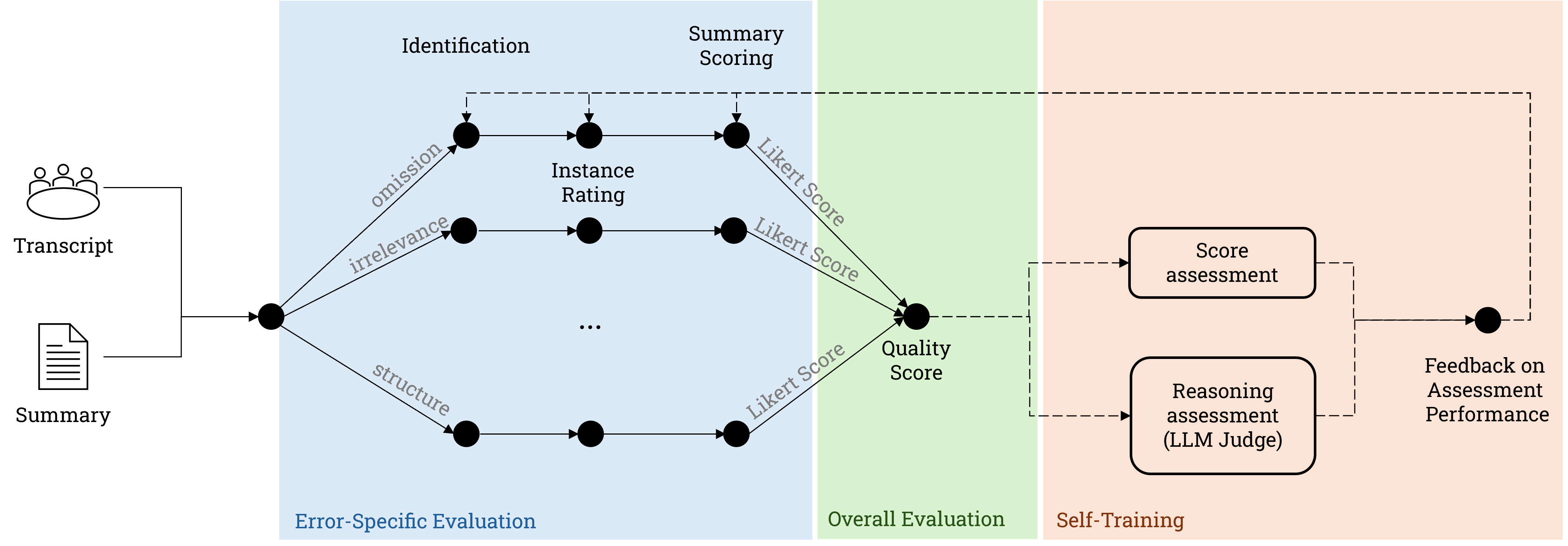}
    \caption{Architecture of \metric{} displaying the singel-aspect assessment using three stages and the self-training mechanic for feedback-based alignment improvement with available human data.}
    \label{fig:main_figure}
\end{figure*}

%% file: text/03_Methodology.tex
Key weaknesses of meeting summarization evaluators include error type confusion \cite{KirsteinRG24a}, oversight of error instances \cite{KirsteinWRG24b}, and risk of self-inconsistency \cite{WeiHXW24}.
To address these, we develop \metric{} through comparative experiments between traditional approaches and promising alternatives.
Our findings indicate that the most reliable, self-consistent, and thorough setup combines error-type specific single-aspect evaluators with multi-agent discussion in a three-stage scoring process (see \Cref{fig:main_figure}).
Experiments use GPT4 backbones, generating verbose confidence scores (0-10) \cite{GengCWK24} and chain-of-thought (CoT) \cite{WeiWSB24} reasoning traces for qualitative analysis.
The prompts and example outputs are provided in \Cref{sec:appendix_prompt,sec:appendix_example}.


\subsection{Error types and dataset}
\label{sec:method_setting}
We assess the error types redundancy (RED), incoherence (INC), language (LAN), omission (OM), coreference (COR), hallucination (HAL), structure (STR), and irrelevance (IRR).
The definitions (see \Cref{sec:appendix_definitions_short}) are based on \citet{KirsteinRG24a}, combining total and partial omission into one.

We use the QMSum Mistake dataset \cite{KirsteinRG24a}, comprising 170 samples from academic (ICSI \cite{JaninBEE03}), business (AMI \cite{MccowanCKA05}), and parliament meetings, summarized by language models (LED \cite{BeltagyPC20}, DialogLED \cite{ZhongLXZ22}, Pegasus-X \cite{PhangZL22}, GPT-3.5, and Phi-3 \cite{AbdinJAA24}) and human-annotated for errors.
Four annotators update the human annotation scores (Likert scale, 0 to 5) and reasoning traces to align with our modified definitions, following the annotation process detailed in \Cref{sec:appendix_annotation}.
We achieve a high inter-annotator agreement of 0.793 (Krippendorff's alpha \cite{Krippendorff70}, complete agreement stated in \Cref{sec:appendix_annotationagreement}), indicating strong reliability.
Statistics on the QMSum Mistake dataset are listed in \Cref{sec:appendix_qmsummistake}.


\subsection{Challenge I: error type confusion}
\label{sec:challenge_1}
Error-type definitions are nuanced (\Cref{sec:appendix_definitions_short}), requiring careful consideration during detection.
Prompting models to consider multiple error types simultaneously (multi-aspect) risks definition confusion \cite{KamoiDLA24}.
Literature suggests restricting detection to one error type at a time (single aspect), using multiple model instances for comprehensive coverage \cite{KirsteinRG24a}.

\paragraph{Single-aspect error-type assessment leads to a more reliable and comprehensive evaluation.}
Multi-aspect approaches often assign uniform scores across error types, provide superficial reasoning (e.g., "it misses details about decision making"), and occasionally confuse error definitions, leading to false detections.
In contrast, single-aspect approaches demonstrate a more thorough understanding of individual error types, identifying a broader range of errors.
However, the single-aspect approach may become oversensitive, assigning overly bad scores to minor errors, aligning with recent findings \cite{KirsteinRG24a}.


\subsection{Chalenge II: error instance oversight}
\label{sec:challenge_2}

A direct assessment of error types may miss critical instances, affecting scoring accuracy \cite{KamoiDLA24}.
We propose a three-step evaluation pipeline to address the risk of oversight and have a more thorough assessment process consisting of identifying potential error instances, rating the error severity for each instance, and assigning a score based on the observations for the currently assessed error type (see \Cref{fig:main_figure}).
Each step is carried out by an LLM instance informed by the result of the previous step. 

\paragraph{Three-step assessment offers more thorough error instance identification and sensitive scoring.}
Comparing single-step and three-step evaluation approaches reveals notable improvements in error detection and scoring with the three-step method.
Using the single-aspect setup as the backbone, we observe that the three-step approach more effectively detects non-obvious error instances, such as paraphrased repetitions. 
Balanced accuracy scores (\Cref{tab:bacc_error_instance_identification_transposed}, definition in \Cref{sec:appendix_b-acc}) show an improvement in detecting all error types with an average improvement of $\sim$3.5\% on average.

However, this increased sensitivity and larger number of detections can lead to overly strict assessments, particularly for subjective error types (e.g., irrelevance).
We conclude that the three-step approach offers a more comprehensive evaluation but requires adjustment, e.g., through in-context samples, to better align with human judgment.
While offering more comprehensive evaluations, the three-step approach requires fine-tuning, potentially through in-context samples, to better align with human judgment.

\input{tables/02_BACC-identeification}


\subsection{Challenge III: inconsistent scoring}
\label{sec:challenge_3}
\input{figures/discussion_paradigms}

To address score fluctuations in LLM-based assessments \cite{WeiHXW24}, we explore a multi-agent debate protocol (MADP) \cite{LiangHJW23}.
In MADP, different models (agents) collaborate through a natural language exchange to solve a task.
We use MADP to challenge and refine an initial draft (e.g., collection of potential error instances).
First, a moderator model provides a draft solution, followed by multiple model instances independently challenging the draft from different perspectives and refining the solution.
Finally, a moderator synthesizes the refinements into a final output.
Through this approach, we embed an additional layer to identify and mitigate false positive or false negative detection, contributing to a more robust and consistent evaluation.

\paragraph{MADP enhances evaluation depth and nuance, improving the overall assessment quality.}
We compare three setups: single-model without MADP (Single), MADP with multiple GPT4 instances (MADP-S), and MADP with diverse models, including GPT4, Phi-3-medium-128k \cite{AbdinJAA24}, Llama 3.2 11b \cite{MetaAI24}, and Gemini 1.5 Flash \cite{GeminiTeamRST24} (MADP-M).
All setups use a single-aspect three-step architecture as base.
Both MADP approaches demonstrate improved error impact sensitivity with more fine-grained explanations and ratings.
The MADP-M offers slightly more diverse perspectives but broadly aligns with MADP-S results. 
\Cref{tab:variance_MAD} shows that score variance can be notably reduced with MADP, with slightly less variance when using only GPT4 instances.

\input{tables/02_Variance_consistency}

\input{tables/03_existance_correlation}

\subsection{Resulting \metric{ } architecture}
\label{sec:metric_setup}
The derived \metric architecture combines single-aspect, three-step evaluation using single-model MADP for thorough assessment. 
Individual error-type Likert scores are combined using a weighted sum, following the idea of \cite{LiuIXW23}:

\begin{equation}
    impact = \frac{\sum_n s_n \cdot (c_n \cdot i_n)}{\sum_n (c_n \cdot i_n)}
\end{equation}

where $s_n$ is the Likert score, $c_n$ the scaled confidence score (0-1) reported by the LLM, and $i_n$ an importance parameter (default: 1.0; OM, HAL, IRR: 1.1; REP, INC, LAN: 0.9).
Errors such as OM, HAL, and IRR are prioritized as they significantly affect summary accuracy and introduce biases, undermining the summary's trustworthiness.
REP, INC, and LAN primarily influence readability and occur less frequently in LLM-generated summaries \cite{KirsteinWRG24b}, warranting a slightly lower weight.
The $impact$ score, describing how large the impact of all errors is on the summary quality (none: 0 to highly impacted: 5), is converted to a quality score (1 to 10) using:

\begin{equation}
    quality = 1 + \left(\frac{5 - impact}{5} \cdot 9 \right)
\end{equation}

An optional \textbf{self-training} mechanism inspired by self-teaching \cite{WangKGY24} and feedback techniques \cite{KirsteinRG24a} is introduced to address overly harsh scoring.
This mechanism uses GPT4 as a judge \cite{ZhengCSW24} to evaluate the quality of the reasoning traces on completeness, overlap with human reasoning, and logic.
For the score differences, we report labels ranging from "no difference" to "major difference" for score discrepancies, with "critical disagreement" for conflicting error observations.
A second GPT4 judge is tasked to detect patterns in the per-sample feedback and provides a consolidated report for each error type on what should be considered or treated differently during evaluation.
This report is then used in the following three-step assessment, being appended to the original task describing prompt to steer the detection and evaluation behavior.

%% file: tables/02_BACC-identeification.tex
\begin{table}[ht]
    \centering
    \tiny
    \begin{tabular}{lcccccccc}
        \toprule
        \textbf{Step} & \textbf{OM} & \textbf{REP} & \textbf{INC} & \textbf{COR} & \textbf{HAL} & \textbf{LAN} & \textbf{STR} & \textbf{IRR} \\
        \midrule
        single  & 93.0  & 93.7  & 88.5  & 85.3  & 71.0  & 85.9  & 87.0  & 81.0 \\
        three   & 95.3  & 94.1 & 90.1  & 89.0  & 77.6  & 90.4  & 89.2  & 87.4 \\
        \bottomrule
    \end{tabular}
    \caption{Balanced accuracy of the error type identification compared against human judgments using the single-step (single) and three-step (three) approach on the modified QMSum Mistake dataset. Error type abbreviations follow the definition in \Cref{sec:appendix_definitions_short}.}
    \label{tab:bacc_error_instance_identification_transposed}
\end{table}

%% file: figures/discussion_paradigms.tex
\begin{figure}
    \centering
    \includegraphics[width=1\linewidth]{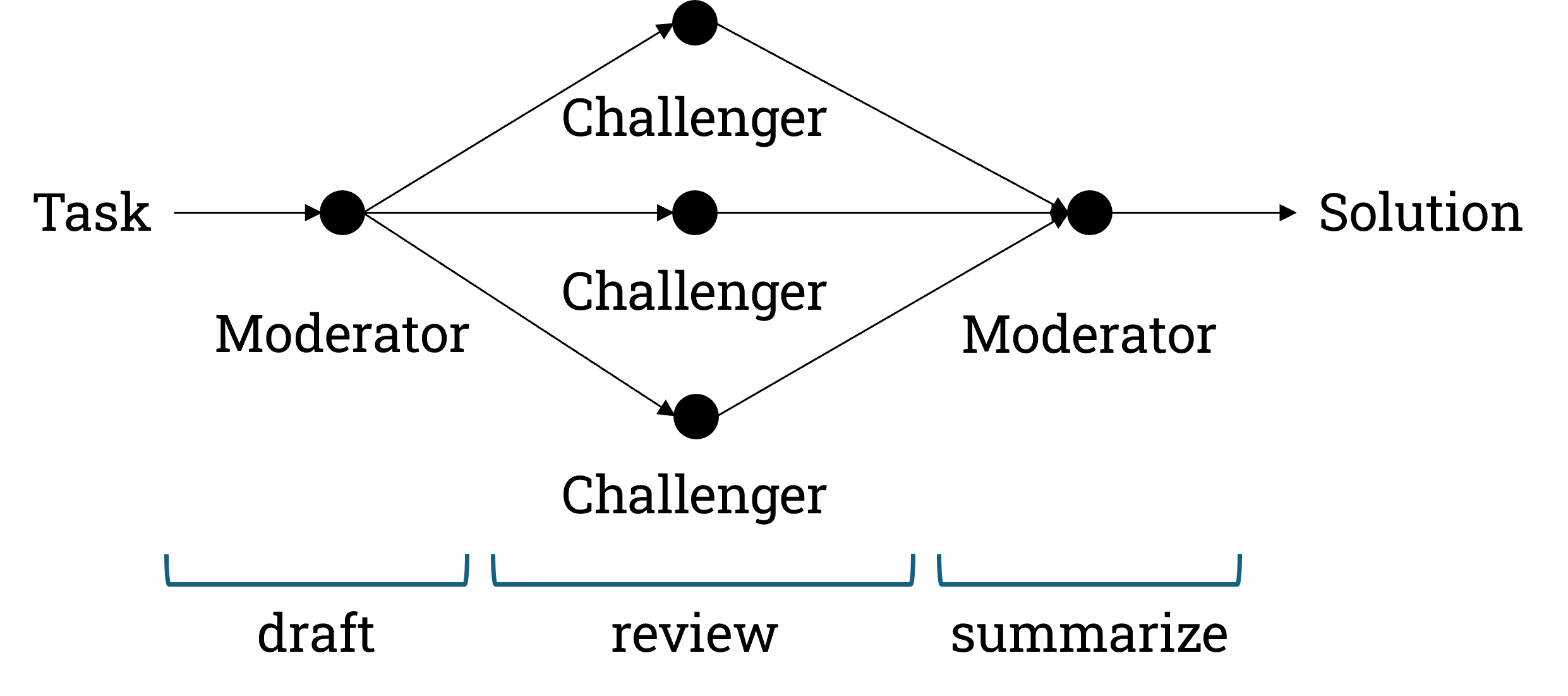}
    \caption{Multi-agent discussion protocol used, consisting of an initial draft generator, three synchronously acting challengers, and a moderator summarizing the individual statements into a final task solution.}
    \label{fig:discussion_paradigms}
\end{figure}

%% file: tables/02_Variance_consistency.tex
\begin{table}[ht]
    \centering
    \tiny
    \setlength{\tabcolsep}{2pt} 
    \renewcommand{\arraystretch}{1.2} 
    \begin{tabular}{l p{0.7cm} p{0.7cm} p{0.7cm} p{0.7cm} p{0.7cm} p{0.7cm} p{0.7cm} p{0.7cm}}
        \toprule
        \textbf{Setup} & \textbf{OM} & \textbf{REP} & \textbf{INC} & \textbf{COR} & \textbf{HAL} & \textbf{LAN} & \textbf{STR} & \textbf{IRR} \\
        \midrule
        \multirow{2}{*}{single}   &  4.08 & 3.74 & 4.03 & 3.39 & 3.81 & 3.76 & 3.83 & 3.38 \\
                                  & (0.01) & (0.07) & (0.07) & (0.26) & (0.29) & (0.06) & (0.11) & (0.08) \\
        \multirow{2}{*}{MADP-S}  &  4.30 & 3.93 & 4.05 & 3.96 & 3.94 & 3.80 & 4.03 & 3.74 \\
                                &  (0.03) & (0.00) & (0.04) & (0.11) & (0.23) & (0.07) & (0.01) & (0.04) \\ 
        \multirow{2}{*}{MADP-M}   &  4.31 & 3.95 & 3.98 & 3.91 & 3.98 & 3.78 & 4.05 & 3.76 \\
                                 &  (0.04) & (0.05) & (0.05) & (0.14) & (0.22) & (0.03) & (0.09) & (0.07) \\
        \bottomrule
    \end{tabular}
    \caption{Mean Likert scores and standard deviation in parentheses below across three iterations. Error type abbreviations follow definition in \Cref{sec:appendix_definitions_short}. Single refers to single LLM setup, MADP-S is MADP with only GPT4 instances, MADP-M is MADP with multi-model instances.}
    \label{tab:variance_MAD}
\end{table}

%% file: tables/03_existance_correlation.tex
\begin{table*}[ht]
    \centering
    \tiny
    \begin{tabular}{lcccccccc}
        \toprule
        \textbf{Step} & \textbf{OM} & \textbf{REP} & \textbf{INC} & \textbf{COR} & \textbf{HAL} & \textbf{LAN} & \textbf{STR} & \textbf{IRR} \\
        \midrule
        ROUGE-1  & 0.01 & 0.13 & -0.02 & 0.06 & 0.13 & 0.02 & 0.09 & -0.23* \\
        ROUGE-2  & -0.00 & 0.20* & 0.08 & 0.15 & 0.15 & 0.12 & 0.17 & -0.11 \\
        ROUGE-LS & 0.07 & 0.26** & 0.08 & 0.19* & 0.19* & 0.05 & 0.23* & -0.20* \\
        BERTScore & -0.10 & -0.04 & -0.15 & 0.08 & 0.01 & -0.24* & 0.08 & -0.32** \\
        \midrule
        G-Eval-4 & -0.13 & -0.49** & -0.24 & -0.21* & -0.26* & -0.21* & -0.21 & -0.16 \\
        \midrule
        Single-0 & -0.25* & -0.48** & -0.39** & -0.22* & -0.14 & -0.23* & -0.35** & -0.12 \\
        Single-1 & -0.26* & -0.53** & -0.42** & -0.25 & -0.27* & -0.28* & -0.41** & -0.13 \\
        Multi-0  & \textbf{-0.30**} & -0.45** & -0.38** & -0.30** & -0.18 & -0.46** & -0.35** & -0.16 \\
        Multi-1  & -0.27** & \textbf{-0.69**} & \textbf{-0.63**} & \textbf{-0.35} & \textbf{-0.33**} & \textbf{-0.52**} & \textbf{-0.43**} & \textbf{-0.21*} \\

        \bottomrule
    \end{tabular}
    \caption{Point-Biserial correlation between metric scores and human annotation. Significant values: * (p $\leq$ 0.05) and ** (p $\leq$ 0.01). Negative correlation means error presence leads to metric score decrease. \textbf{Bold} means best value.}
    \label{tab:error_existence_correlation}
\end{table*}

%% file: text/04_Experiments.tex
\subsection{Setup}

We compare \metric{} with established metrics using the modified QMSum Mistake dataset and the eight error types: omission (OM), repetition (REP), incoherence (INC), coreference (COR), hallucination (HAL), language (LAN), structure (STR), and irrelevance (IRR). 
We use the \metric{} setup described in \Cref{sec:metric_setup} with and without MADP (Multi-n, Single-n), with n iterations of self-training (0, 1).

Baseline metrics include:
\begin{itemize}
    \item \textit{ROUGE} \cite{Lin04}, the most common, count-based metric, assessing n-gram overlap between generated and reference summaries. We report unigrams, bigrams, and the longest common sequence.
    \item \textit{BERTScore} \cite{ZhangKWW20}, a model-based metric measuring the contextual similarity between generated and reference texts, reflecting semantic and syntactic similarity. We report the rescaled F score\footnote{\href{https://github.com/Tiiiger/bert_score/blob/master/journal/rescale_baseline.md}{https://github.com/Tiiiger/bert score/blob/master/journal/rescale baseline.md}}.
    \item A modified version of the LLM-based \textit{G-Eval-4} \cite{LiuIXW23} prompted with our eight evaluation criteria and access to the transcript.
\end{itemize}


\input{tables/03_sensitivity_correlation}

\subsection{Analysis and discussion}

Our analysis focuses on three aspects of evaluation: error masking, sensitivity to error impact, and closeness to human ratings.
We conclude that the three-stage detection in \metric{} demonstrates significant improvements over the best current approach, G-Eval-4, showing the highest correlation with human judgment on both pure error detection (avg. gap: 0.1) and error sensitivity (avg. gap: 0.15).
The self-teaching loop further enhances \metric{}'s performance, increasing correlation (avg. gap increase: 0.1) and notably closing the gap to human judgment (up to 1.4 points reduction).
Multi-1 exhibits the best assessment performance, while Single-1 offers a faster, less computationally expensive alternative with a slight performance decrease. 

\paragraph{\metric{} demonstrates a high correlation on error existence, indicating a low error masking tendency.}
\Cref{tab:error_existence_correlation} shows the Point-Biserial correlation \cite{Tate54} analysis between considered automatic metrics and human annotation.
Traditional count- and model-based metrics (ROUGE, BERTScore) perform poorly across most dimensions as expected \cite{KirsteinWRG24b}.
LLM-based methods show higher, desired negative correlations with human judgment, suggesting them as a preferred choice.
G-Eval-4 exhibits mostly weak correlations, with stronger reactions for REP, INC, and STR.
We hypothesize that not all error instances are detected by G-Eval-4, leading to erroneous evaluation behavior.

\metric's Multi-n and Single-n setups surpass previous state-of-the-art evaluators in correlation across all error types (avg. -0.13 compared to G-Eval-4), indicating the benefit of splitting assessment into dedicated detection and scoring.
INC, LANG, and IRR benefit most, while OM and HAL remain challenging, aligning with recent findings on LLMs' struggle with contextualization \cite{KirsteinRG24a}.
As qualitative analysis reveals, self-training further provides a slight boost by asking the model to prioritize identified error instances explicitly.
MADP-based variants achieve greater correlation, indicating that the refinement process helps eliminate falsely detected instances and consider overlooked ones.

\input{tables/03_gap}

\paragraph{\metric{}'s rating of individual error instances helps capture error type severity in scores.}
\Cref{tab:error_sensitivity_correlation} shows Kendall \cite{Kendall38} and Spearman \cite{Spearman04} correlations between automatic metrics and human annotations on error type impact.
ROUGE and BERTScore correlate well for IRR errors but struggle elsewhere, with BERTScore rewarding severe REP instances and ROUGE tending to reward HAL.
LLM-based metrics demonstrate weak to mid-negative correlations, indicating a capability to understand and reflect varying impact severities in score.

\metric's multi-step approach outperforms current methods, suggesting that previous limitations may stem from overlooking score-influencing error instances, leading to a weaker reflection of error impacts in scores.

The improvement through MADP indicates that reflective discussion enhances the categorization of error instance impacts and promotes a more thorough score reassessment.
Self-training further boosts performance (average improvement of -0.1), demonstrating that feedback on reasoning traces and scoring behavior aids in error categorization.


\paragraph{Self-teaching addresses the initial overestimation of error impact.}
\Cref{tab:gap} shows that the gap between \metric-assigned and human-annotated Likert scores is initially greater than for LLM-based metrics relying on a single-step assessment.
This greater gap may be due to the more thorough error detection with the three-step assessment pipeline, leading the framework to assign higher scores than humans.
However, self-teaching feedback drastically narrows this gap by up to 1.4 points, lowering it below baseline gaps.

%% file: tables/03_sensitivity_correlation.tex
\begin{table*}[ht]
    \centering
    \tiny
    \setlength{\tabcolsep}{2pt} 
    \begin{tabular}{lcccccccccccccccc}
        \toprule
        & \multicolumn{2}{c}{\textbf{OM}} & \multicolumn{2}{c}{\textbf{REP}} & \multicolumn{2}{c}{\textbf{INC}} & \multicolumn{2}{c}{\textbf{COR}} & \multicolumn{2}{c}{\textbf{HAL}} & \multicolumn{2}{c}{\textbf{LAN}} & \multicolumn{2}{c}{\textbf{STR}} & \multicolumn{2}{c}{\textbf{IRR}} \\
        \cmidrule(lr){2-3} \cmidrule(lr){4-5} \cmidrule(lr){6-7} \cmidrule(lr){8-9} \cmidrule(lr){10-11} \cmidrule(lr){12-13} \cmidrule(lr){14-15} \cmidrule(lr){16-17}
        \textbf{Step} & $\rho$ & $\tau$ & $\rho$ & $\tau$ & $\rho$ & $\tau$ & $\rho$ & $\tau$ & $\rho$ & $\tau$ & $\rho$ & $\tau$ & $\rho$ & $\tau$ & $\rho$ & $\tau$ \\
        \midrule
        ROUGE-1  
        & -0.03 & -0.03 & 0.11 & 0.08 & 0.00 & 0.00 & 0.08 & 0.06 & 0.22* & 0.15* & 0.01 & 0.01 & 0.08 & 0.07 & -0.24** & -0.18** \\
        ROUGE-2  
        & -0.03 & -0.02 & 0.16 & 0.12 & 0.03 & 0.03 & 0.12 & 0.10 & 0.18* & 0.13 & 0.06 & 0.05 & 0.12 & 0.10 & -0.15 & -0.11 \\
        ROUGE-LS 
        & -0.06 & -0.04 & 0.10 & 0.07 & 0.03 & 0.02 & 0.07 & 0.06 & 0.18 & 0.13 & -0.01 & -0.01 & 0.06 & 0.05 & -0.21* & -0.16* \\
        BERTScore 
        & 0.07 & -0.01 & 0.22* & 0.17* & -0.20* & -0.15* & 0.03 & 0.02 & -0.05 & 0.04 & 0.05 & 0.02 & 0.02 & 0.02 & \textbf{-0.44**} & \textbf{-0.34**} \\
        \midrule
        G-Eval-4
        & -0.24* & -0.18* & -0.44** & -0.34** & -0.36** & -0.28** & -0.15 & -0.12 & -0.18* & -0.14* & -0.22* & -0.18* & -0.15 & -0.13 & -0.17 & -0.13 \\
        \midrule
        Single-0
        & -0.27* & -0.20* & -0.47** & -0.36** & -0.42** & -0.32** & -0.24* & -0.19* & -0.22* & -0.16* & -0.25* & -0.19* & -0.37** & -0.29** & -0.22* & -0.16* \\
        Single-1
        & -0.42** & -0.32** & -0.53** & -0.41** & -0.46** & -0.35** & -0.27** & -0.22** & \textbf{-0.26*} & \textbf{-0.19*} & -0.30* & -0.23** & \textbf{-0.40**} & \textbf{-0.31**} & -0.21* & -0.16* \\
        Multi-0
        & -0.31 & -0.22 & -0.52** & -0.41** & -0.34 & -0.24 & \textbf{-0.35*} & \textbf{-0.29*} & -0.19 & -0.13 & \textbf{-0.49**} & -0.37** & -0.34** & -0.27** & -0.25 & -0.20 \\
        Multi-1
        & \textbf{-0.58**} & \textbf{-0.46**} & \textbf{-0.57**} & \textbf{-0.46**} & \textbf{-0.58**} & \textbf{-0.45**} & -0.33** & -0.27** & -0.22* & -0.16* & \textbf{-0.49**} & \textbf{-0.40**} & -0.37** & -0.29** & -0.34** & -0.26** \\
        
        \bottomrule
    \end{tabular}
    \caption{Kendall ($\tau$) and Spearman ($\rho$) correlation between metric scores and human annotation. Significant values: * (p $\leq$ 0.05) and ** (p $\leq$ 0.01). Negative correlation: high impact leads to metric score decrease. \textbf{Bold}: best value.}
    \label{tab:error_sensitivity_correlation}
\end{table*}

%% file: tables/03_gap.tex
\begin{table}[ht]
    \centering
    \tiny
    \begin{tabular}{lcccccccc}
        \toprule
        \textbf{Step} & \textbf{OM} & \textbf{REP} & \textbf{INC} & \textbf{COR} & \textbf{HAL} & \textbf{LAN} & \textbf{STR} & \textbf{IRR} \\
        \midrule
        G-Eval-4  & 0.56  & 1.97  & 2.30  & 2.60  & 1.10  & 2.07  & 2.53  & 1.68 \\
        \midrule
        Single-0  & 0.73  & 2.36  & 2.92  & 2.77  & 1.50  & 2.73  & 2.79  & 1.91 \\
        Single-1  & 0.31  & 1.87  & 2.15  & 2.70  & 1.17  & 2.02  & 2.34  & 1.70 \\
        \midrule
        Multi-0  & 0.92  & 2.60  & 2.96  & 3.24  & 2.03  & 2.87  & 3.06  & 2.39 \\
        Multi-1  & 0.22  & 1.71  & 1.53  & 2.46  & 1.06  & 2.13  & 2.33  & 1.83 \\
        \bottomrule
    \end{tabular}
    \caption{Gap of the mean LLM-assigned Likert scores to the mean human-assigned Likert scores for the individual error types.}
    \label{tab:gap}
\end{table}

%% file: text/02_Related_Work.tex
\noindent \textbf{Meeting summarization evaluation}  faces significant challenges with traditional metrics like ROUGE \cite{Lin04} and BERTScore \cite{ZhangKWW20}.
These metrics correlate relatively poorly with human judgment, potentially masking or rewarding certain error types (e.g., QuestEval \cite{ScialomDLP21} favors missing information).
LLM-generated summaries expose these limitations further, leading to minimal metric score differences despite substantial qualitative variations \cite{KirsteinRG24a}.
Our work formalizes the error-type focused evaluation concepts by \citet{KirsteinRG24a} into a thorough detection framework.

\noindent \textbf{LLMs as summary evaluators} have shown promising results, with approaches like GPTScore \cite{FuNJL24}, G-EVAL \cite{LiuIXW23}, and self-taught evaluators \cite{WangKGY24} demonstrating positive correlation with human judgments.
For meeting summarization specifically, single-evaluator metrics such as AUTOCALIBRATE \cite{LiuYHZ23} and FACTSCORE \cite{MinKLL23} are recently explored but still lag in reliability and alignment with human judgment \cite{KirsteinRG24a}.
Persistent challenges include difficulty detecting specific error types (e.g., omission) and handling subjective assessments \cite{KirsteinRG24a}.
Our work continues research of LLM-based metrics by further developing existing objective error definitions \cite{KirsteinRG24a}, implementing an LLM-based single-aspect evaluator, and incorporating a refinement process inspired by the self-teaching technique \cite{WangKGY24}.

%% file: text/B_Appendix_Prompts.tex
\label{sec:appendix_prompt}
In the following, we present the prompts used to identify errors (\Cref{fig:step_1_identification}), rate the severity of these instances (\Cref{fig:step_2_rating}), and to assign the impact score (\Cref{fig:step_3_scoring}).

\begin{figure*}[t]
    \begin{AIbox}{Step 1: Error Instance Identification Prompt Template}
    \parbox[t]{\textwidth}{
        Step 1 is to collect possible error instances.
        \newline
        Read the following criteria carefully: *** {criteria}: {self.criteria[criteria]} ***.
        \newline
        Next, read the summary: ** {data['summary']} **.
        \newline
        Also, consider the original meeting transcript: * {data['transcript']} *. \newline
        \newline
        Now, read the summary again and write down a list of instances where this error type could occur. This can contain instances that already show the error or instances that could potentially show the error.
        For every instance, write down a short reasoning thinking step-by-step why this instance could be an error.
        Also, for every instance, provide a score from 0 (totally unsure) to 100 (totally sure) to show how certain you are that this instance could be an error.
        Ensure that each instance is provided in strict JSON format, using double quotes for keys and values, and no additional text outside the JSON structure.
        Return your answer only in the following format:\newline
        \newline
        [ \{'instance': '<text passage or sentence or words from summary>', 'reasoning' : '<chain-of-thought reasoning>', 'certainty': '<score from 0 meaning totally unsure to 100 meaning totally sure>'\}, \{<same for instance 2>\}, ... \{<same for instance n>\}]\newline
        \newline
        Ensure that the format strictly follows valid JSON, with no extra preambles or additional information.
    }
    \end{AIbox}
    \caption{The prompt template used to task an LLM instance to identify potential error instances.}
    \label{fig:step_1_identification}
\end{figure*}

\begin{figure*}[t]
    \begin{AIbox}{Step 2: Error Instance Rating Prompt Template}
    \parbox[t]{\textwidth}{
        Step 2 is to rate the severity of the potential error instances.\newline
        Read the following criteria carefully: *** {criteria}: {self.criteria[criteria]} ***.\newline
        Next, read the already collected potential error instances: *** {list\_of\_instances} ***.\newline
        Also, consider the original meeting transcript: * \{data['transcript']\} * and the summary: * {data['summary']} *.\newline
        \newline
        Now, for each instance, decide if it is an actual error instance or not according to the criteria.
        For each instance, write down a short reasoning explaining why you decided so.
        Provide a score on the severity of the error, ranging from 0 (no error) to 10 (severe error).
        Also, provide a score for your certainty, ranging from 1 (totally unsure) to 10 (totally sure).
        For each instance, indicate whether the error exists by setting the 'error\_exists' field to true or false.
        Return the output strictly in JSON format, using double quotes around all keys and values, and return nothing else.
        Here is the required format for your response: \newline
        \newline
        [ \{'instance': '<the instance>', 'reasoning' : '<chain-of-thought reasoning if there is an error according to the criteria or not>', 'certainty': '<score from 0 meaning totally unsure to 100 meaning totally sure>', 'error\_exists' : <true or false depending on your decision>\}, \{<same for instance 2>\}, ... \{<same for instance n>\}]\newline
        \newline
        Make sure the output is strictly valid JSON, with no preamble, extra explanations, or text outside the JSON structure.
    }
    \end{AIbox}
    \caption{The prompt template used to task an LLM instance to rate detected error instance.}
    \label{fig:step_2_rating}
\end{figure*}

\begin{figure*}[t]
    \begin{AIbox}{Step 3: Scoring Prompt Template}
    \parbox[t]{\textwidth}{
        Step 3 is to rate the summary considering the actual error instances and their severity. \newline
        Read the following criteria carefully: *** {criteria}: {self.criteria[criteria]} ***. \newline
        Consider the observed error instances and their severity scores (0 for no error to 10 meaning severe error): *** {list\_of\_instances} ***. \newline
        You do not have to agree with these severity scores, so please critically evaluate them when rating the summary.
        Next, read the summary: ** {data['summary']} **. \newline
        Consider the original meeting transcript: * {data['transcript']} *. \newline
        \newline
        Now, rate the summary with a single score from 0 to 5, where 0 means no impact at all (a really good summary) and 5 means a very high impact (a poor summary) regarding this error type.
        Also, provide a short reasoning explaining why you rated the summary as you did.
        Additionally, provide a certainty score indicating how confident you are in your rating, from 0 (not confident at all) to 10 (very confident).
        Return your answer strictly in **valid JSON format**, using **double quotes** around all keys and string values. \newline
        The structure of your response should be exactly like this: \newline
        '\{ "reasoning": "<your reasoning>", "confidence": <your certainty score>, "rating": <score from 0 to 5> \} \newline
        Make sure to return only the JSON response, without any preambles, explanations, or extra text outside the JSON format.  
    }
    \end{AIbox}
    \caption{The prompt template used to task an LLM instance to perform the scoring.}
    \label{fig:step_3_scoring}
\end{figure*}

%% file: text/C_Appendix_ExampleOutputs.tex
\label{sec:appendix_example}

In \Cref{tab:single_vs_multi_aspect}, we show the output differences between the multi- and single-aspect setups from \Cref{sec:challenge_1}.
\Cref{tab:single-step_vs_three-step} shows the difference when using the three-split identification and assessment approach detailed in \Cref{sec:challenge_2}.
The influence of MADP (\Cref{sec:challenge_3}) and the usage of a single or multiple model families is shown in \Cref{tab:MADP}.

\begin{table*}[h]
    \centering
    \small
    \begin{tabular}{lp{12cm}}
    \toprule
        \textbf{Setup}       & \textbf{Evaluator Output} \\
    \midrule
    multi-aspect   &  The summary captures the main points of the discussion, such as the challenges in creating a database format, the pros and cons of different formats, and the need for flexibility and ease of use. However, it omits specific details about the decisions made, the exact formats discussed (like ATLAS and P-files), and the nuances of the arguments presented. This leads to an incomplete understanding of the meeting's outcomes.\\
      \midrule
    single-aspect   &  The summary captures the main points of the discussion, including the challenges of creating a database format for linking transcriptions, the pros and cons of different formats, and the need for compactness, ease of use, and flexibility. However, it omits some specific details such as the mention of specific tools like ATLAS, and the discussion about the potential use of relational databases. These omissions slightly reduce the comprehensiveness of the summary.\\
    \bottomrule
    \end{tabular}
\caption{Outputted reasoning of the multi- and single-aspect setups shown in \Cref{sec:challenge_1}. Single-aspect is taken from the omission error type while multi-aspect considerat all error types.}
\label{tab:single_vs_multi_aspect}
\end{table*}

\begin{table*}[h]
    \centering
    \small
    \begin{tabular}{lp{12cm}}
    \toprule
        \textbf{Setup}       & \textbf{Evaluator Output} \\
    \midrule
    single-step   &  The summary captures the main points of the discussion, including the challenges of creating a database format for linking transcriptions, the pros and cons of different formats, and the need for compactness, ease of use, and flexibility. However, it omits some specific details such as the mention of specific tools like ATLAS, and the discussion about the potential use of relational databases. These omissions slightly reduce the comprehensiveness of the summary.\\
      \midrule
    three-step   &  The summary omits critical details about the specific pros and cons of different formats discussed, the detailed arguments about how each format meets the needs of compactness, ease of use, and handling multiple speakers, the compatibility and features of NIST's format, the long-term implications and flexibility concerns, and the specific challenges and considerations in creating a database format. These omissions significantly impact the understanding of the decision-making process and the depth of the group's concerns.\\
    \bottomrule
    \end{tabular}
\caption{Outputted reasoning of the single-step and three-step setups shown in \Cref{sec:challenge_2} on the omission error type.}
\label{tab:single-step_vs_three-step}
\end{table*}

\begin{table*}[h]
    \centering
    \small
    \begin{tabular}{lp{12cm}}
    \toprule
        \textbf{Setup}       & \textbf{Evaluator Output} \\
    \midrule
    no MADP   &  The summary omits critical details about the specific pros and cons of different formats discussed, the detailed arguments about how each format meets the needs of compactness, ease of use, and handling multiple speakers, the compatibility and features of NIST's format, the long-term implications and flexibility concerns, and the specific challenges and considerations in creating a database format. These omissions significantly impact the understanding of the decision-making process and the depth of the group's concerns.\\
      \midrule
    MADP + single   &  The summary omits several critical details discussed in the meeting, such as the specific pros and cons of different formats (XML, P-files, Transcriber), the detailed comparison with NIST's format, and the specific challenges and considerations for long-term implications and flexibility. These omissions lead to an incomplete understanding of the meeting's discussions and decisions, which could result in misinformed actions based on the summary. For instance, the lack of detail on the compactness of P-files versus XML, or the handling of multiple speakers, could lead to choosing an inefficient format. Similarly, not mentioning the potential need for translation between formats or the scalability issues could result in future technical debt. The severity of these omissions is high because they directly affect the core objectives of the meeting, which is to decide on a suitable database format.\\
    MADP + multi   &  The summary omits several critical aspects of the discussions, such as the detailed pros and cons of different formats (XML, P-files, Transcriber format), specific technical details and arguments for and against each format, the detailed comparison between NIST's ATLAS format and other formats, and specific challenges like issues with frame-level data, the need for binary encoding for large datasets, and potential problems with merging different annotations. These omissions lead to an incomplete representation of the meeting, which could result in misinformed decisions or actions based on the summary.\\
    \bottomrule
    \end{tabular}
\caption{Outputted reasoning of the additional usage of MADP with onle a single backbone model (MADP + single) or models from different model families (MADP + multi), as described in \Cref{sec:challenge_3}.}
\label{tab:MADP}
\end{table*}

%% file: text/D_Appendix_ErrorTypeDefinitions.tex
\label{sec:appendix_definitions_short}

We show the short error type definitions in \Cref{tab:error_definition}.
The full-length definitions used for prompting will be made available in the project accompanying GitHub repository.

\begin{table*}
\centering
\small
\begin{tabular}{p{2cm} p{13cm}}
\toprule
\textbf{Error Type} & \textbf{Definition} \\
\midrule
Redundancy \newline \textbf{RED}  & The summary contains repeated or redundant information, which does not help the understanding or contextualization. \\
\midrule
Incoherence \newline \textbf{INC}  & The model generates summaries containing characteristics that disrupt the logical flow, relevance, or clarity of content either within a sentence (intra-sentence) or across sentences (inter-sentence). \\
\midrule
Language \newline \textbf{LAN}  & The model uses inappropriate, incorrect (ungrammatical), or ambiguous language or fails to capture unique linguistic styles. \\
\midrule
Omission \newline (partial, total) \newline \textbf{P-OM}, \textbf{T-OM}  & Missing information from the meeting, such as significant decisions or actions. \textbf{Total omission:} Relevant topics and key points are not stated. \textbf{Partial omission:} Salient topics are mentioned but not captured in detail. \\
\midrule
Coreference \newline \textbf{COR}  & The model fails to resolve a reference to a participant or entity, misattributes statements, or omits necessary mentions. \\
\midrule
Hallucination \newline \textbf{HAL}  & The model produces inconsistencies not aligned with the meeting content. \textbf{Intrinsic:} Misrepresents information from the transcript. \textbf{Extrinsic:} Introduces content not present in the transcript. \\
\midrule
Structure \newline \textbf{STR}  & The model misrepresents the order or logic of the meeting's discourse, misplacing topics or events. \\
\midrule
Irrelevance \newline \textbf{IRR}  & The summary includes information that is unrelated or not central to the main topics or objectives of the meeting. \\
\bottomrule
\end{tabular}
\caption{Definition of the eight error types annotated in QMSum Mistake based on existing error types \cite{KirsteinRG24a,ChangLGI24}}
\label{tab:error_definition}
\end{table*}


%% file: text/E_Appendix_Dataset.tex
\subsection{QMSum Mistake Statistics}
\label{sec:appendix_qmsummistake}

In \Cref{tab:statistics_QMSum} we show the statistics of our modified QMSum Mistake variant.

\begin{table*}
  \small
  \centering
    \begin{tabular}{lcccccc}
    \toprule
    Dataset & \# Meetings & \# Turns & \# Speakers & \# Len. of Meet. & \# Len. of Gold Sum. & \# Len. of Aut. Sum. \\
    \midrule
    QMSum Mistake & 200 (169)   & 556.8 & 9.2   & 9069.8 & 109.1 & 116.9 \\
    \bottomrule
    \end{tabular}%
  \caption{Statistics for the QMSum Mistake dataset. Values are averages of the respective categories. Lengths (Len.) are in number of words. In \# Meetings, values in parentheses are the number of erroneous samples.}
  \label{tab:statistics_QMSum}%
\end{table*}%

\subsection{Annotation Process}
\label{sec:appendix_annotation}

\paragraph{Annotator selection:}
Our annotation team consisted of four graduate students, officially employed as interns or doctoral candidates through standardized contracts.
We selected them from a pool of volunteers based on their availability to complete the task without time pressure and their English proficiency (native speakers or C1-C2 certified). 
By that, we ensured they could comprehend meeting transcripts, human-written gold summaries from QMSum, and all model-generated summaries.
We aimed for gender balance (1 male, 3 female) and diverse backgrounds, resulting in a team of one computer science student, two psychology students, and one communication science student, aged 22-28.

\paragraph{Preparation:}
We prepared a comprehensive handbook for our annotators, detailing the project context and defining challenges and error types (a short version as presented in Section 3 and a long version with more details).
Each definition included two examples: one with minimal impact (e.g., slight information redundancy) and one with high impact (e.g., repeated information throughout).
The handbook explained the binary yes/no rating for the existence of an error. 
Annotators were further tasked to provide reasoning for each decision. 
The handbook did not specify an order for processing errors.
We provided the handbook in English and in the annotators' native languages, using professional translations.

We further elaborated a three-week timeline for the annotation process, preceded by a one-week onboarding period. The first week featured twice-weekly check-ins with annotators, which were reduced to weekly meetings for the following two weeks. Separate quality checks without the annotators were scheduled weekly.
(Note: week refers to a regular working week)

\paragraph{Onboarding:}
The onboarding week was dedicated to getting to know the project and familiarization with the definitions and data.
We began with a kick-off meeting to introduce the project and explain the handbook, particularly focusing on each definition.
We noted initial questions to potentially revise the handbook. 
Annotators were provided with 35 samples generated by SLED+BART \cite{IvgiSB22}, chosen for their balance of identifiable errors and good-quality summaries while capable of processing the whole meeting.
After the first 15 samples, we held individual meetings to clarify any confusion and updated the guidelines accordingly, mainly focusing on our new omission definition.
The remaining 20 samples were then annotated using these updated guidelines. 
A second group meeting this week addressed any new issues with definitions.
We then met individually with annotators after the group meeting to review their work, ensuring quality and understanding of the task and samples. 
All four annotators demonstrated reliable performance and good comprehension of the task and definitions judging from the reasoning they provided for each decision and annotation.
We computed an inter-annotator agreement score using Krippendorff's alpha, achieving 0.793, indicating sufficiently high overlap.

\paragraph{Annotation Process:}
Each week, we distribute all samples generated by one model/source (on average 33 samples) to one of the annotators. 
Consequently, one annotator worked through all samples of one model/source in one week.
On average, one annotator processes summaries from three models/sources (depending on other commitments, some annotators could only annotate two datasets, and others four or more).
Each sample is annotated by three annotators. 
Annotators were unaware of the summary-generating model and were given a week to complete their set at their own pace and break times. 
Quiet working rooms were provided if needed for concentration.
To mitigate position bias, the sample order was randomized for each annotator. 
Annotators could choose their annotation order for each sample and were allowed to revisit previous samples.
To simplify the process, we framed each error type as a question, such as "Does the summary contain repetition?".

Regular meetings were held to address any emerging issues or questions on definitions. 
During the quality checks performed by the authors, we looked for incomplete annotations, missing explanations, and signs of misunderstanding judging from the provided reasoning. 
In case we would have found such a quality lack, the respective annotator would have been notified to re-do the annotation.
After the three-week period, we computed inter-annotator agreement scores on the error types (shown in \Cref{tab:krippendorffs_alpha_human_annotation}). 
In case we had observed a significant difference across annotators, we had planned a dedicated meeting to discuss such cases with all annotators and a senior annotator. 
On average, annotators spent 37 minutes per sample, completing about 7 samples daily.

\paragraph{Handling of unexpected cases:}
Given that our annotators had other commitments, we anticipated potential scheduling conflicts. 
We allowed flexibility for annotators to complete their samples beyond the week limit if needed, reserving a fourth week as a buffer. 
Despite these provisions, all annotators successfully completed their assigned samples within the original weekly timeframes. 
We further allowed faster annotators to continue with an additional sample set. This additional work was voluntary.

\subsection{Inter annotator agreement}
\label{sec:appendix_annotationagreement}
\Cref{tab:krippendorffs_alpha_human_annotation} shows the inter-annotator agreement scores (Krippendorff's alpha) for our modified version of QMSum Mistake.

\begin{table*} [ht]
  \centering
  \small
  \begin{tabular}{lc}
    \toprule
    \textbf{Assessed Characteristic} & \textbf{Krippendorff's $\alpha$} \\
    \midrule
    Omission   & 0.832  \\
    Repetition & 0.811  \\
    Incoherence     & 0.824   \\
    Coreference     & 0.793   \\
    Hallucination     & 0.820  \\
    Language   & 0.725  \\
    Structure  & 0.745   \\
    Irrelevance & 0.793 \\
    \bottomrule
  \end{tabular}
  \caption{Inter-rater reliability for the human annotations, measured by Krippendorff's alpha. Scores $\geq$ 0.667 mean moderate agreement and scores $\geq$ 0.8 mean strong agreement.}
  \label{tab:krippendorffs_alpha_human_annotation}
\end{table*}

%% file: text/F_Appendix_BACC.tex
\label{sec:appendix_b-acc}
Accuracy (ACC) is a natural choice to measure the proportion of correctly predicted labels out of the total number of labels:

\begin{equation}
    ACC = \frac{(TP + TN)}{(TP + FN + FP + TN)}
\end{equation}

with
\begin{itemize}
    \item TP - true positive
    \item TN - true negative
    \item FP - false positive
    \item FN - false negative
\end{itemize}

In our scenario for assessing the error identification capabilities, accuracy itself is not suitable, as some error types have a notable data imbalance, e.g., omission errors.
Therefore, we report the balanced accuracy (B-ACC), i.e., the arithmetic mean of sensitivity (SEN) and specificity (SPE):

\begin{equation}
    SEN = \frac{TP}{(TP + FN}
\end{equation}

\begin{equation}
    SPE = \frac{TN}{(TN + FP)}
\end{equation}

\begin{equation}
    \text{B-ACC} = \frac{1}{2} (SEN + SPE)
\end{equation}

%% file: annotation.tex
\hypertarget{annotation}{}
\citationtitle

\begin{bibtexannotation}
@inproceedings{kirstein-etal-2025-meetingevaluator,
 author={Kirstein, Frederic and Ruas, Terry and Gipp, Bela},
 title={Is my Meeting Summary Good? Estimating Quality with a Multi-LLM Evaluator},
 booktitle={Proceedings of the 31th International Conference on Computational Linguistics: Industry Track},
 pages={14},
 publisher={International Committee on Computational Linguistics},
 year={2025},
 month={01}
}\end{bibtexannotation}